\documentclass[10pt,twocolumn,letterpaper]{article}

\usepackage{cvpr}              

\usepackage{colortbl}      
\usepackage[table]{xcolor} 
\usepackage{multirow}
\usepackage{comment}
\usepackage{pifont}

\newcommand{\cmark}{\ding{51}}  
\newcommand{\xmark}{\ding{55}}  

\definecolor{customorange}{RGB}{220,120,0} 
\definecolor{cvprblue}{rgb}{0.21,0.49,0.74}
\usepackage[pagebackref,breaklinks,colorlinks,allcolors=cvprblue]{hyperref}

\definecolor{bestcolor}{RGB}{220,242,220}      
\definecolor{secondcolor}{RGB}{236,220,250}    

\def\paperID{} 
\def\confName{CVPR}
\def\confYear{2026}
\def\ours{NAS3R}

\title{From None to All: Self-Supervised 3D Reconstruction via Novel View Synthesis } 

\author{Ranran Huang, Weixun Luo, Ye Mao, Krystian Mikolajczyk \\
Imperial College London\\
{\tt\small \{r.huang24, weixun.luo19, ye.mao21, k.mikolajczyk\}@imperial.ac.uk 
}
}

\begin{document}

\twocolumn[{
\renewcommand\twocolumn[1][]{#1}
\maketitle
\begin{center}
    \centering
    \includegraphics[width=1.0\linewidth,page=1]{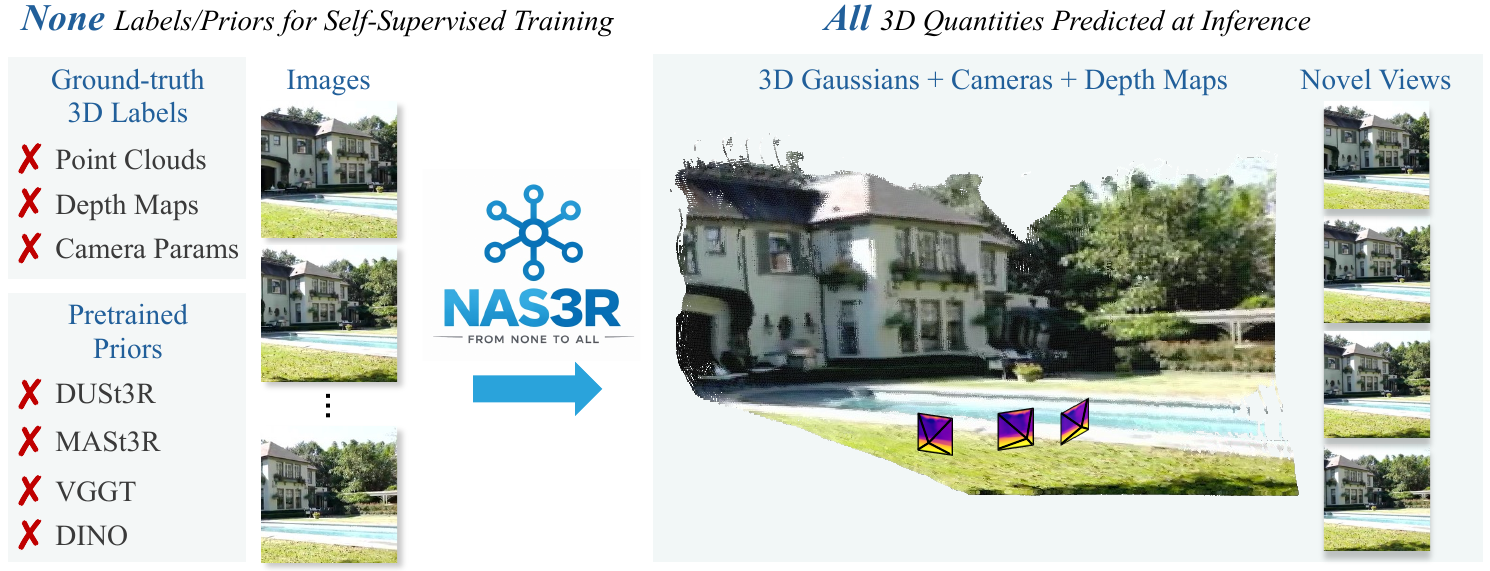}
    \vspace{-2em}
    \captionsetup{hypcap=false} 
    \captionof{figure}{\textbf{\ours} is a self-supervised framework that requires no ground-truth annotations and no pretrained priors during training, and jointly infers 3D Gassian parameters, camera intrinsics and extrinsics and depth maps, also enabling high-quality novel view synthesis.  }
    \label{fig:teaser}
\end{center}
}]
\begin{abstract}

In this paper, we introduce NAS3R, a self-supervised feed-forward framework that jointly learns explicit 3D geometry and camera parameters with no ground-truth annotations and no pretrained priors. During training, NAS3R reconstructs 3D Gaussians from uncalibrated and unposed context views and renders target views using its self-predicted camera parameters, enabling self-supervised training from 2D photometric supervision. To ensure stable convergence, NAS3R integrates reconstruction and camera prediction within a shared transformer backbone regulated by masked attention, and adopts a depth-based Gaussian formulation that facilitates well-conditioned optimization. The framework is compatible with state-of-the-art supervised 3D reconstruction architectures and can incorporate pretrained priors or intrinsic information when available. Extensive experiments show that NAS3R achieves superior results to other self-supervised methods, establishing a scalable and geometry-aware paradigm for 3D reconstruction from unconstrained data.
Code and models are publicly available at \href{https://ranrhuang.github.io/nas3r/}{https://ranrhuang.github.io/nas3r/}.
\end{abstract}    

\vspace{-1.5em}
\section{Introduction}
\label{sec:intro}
Simultaneously recovering 3D structure from 2D images and estimating camera parameters is one of the long-standing and fundamental end-goals in computer vision~\cite{hartley2003multiple, schonberger2016structure, wang2025vggt}. A key challenge is the chicken-and-egg optimization problem~\cite{lin2021barf}: reliable reconstruction requires accurate camera parameters, while camera estimation, in turn, depends on consistent geometric correspondences from the reconstruction. 
While classical methods typically address this problem by utilizing sequential~\cite{lowe2004distinctive,rublee2011orb} and complex iterative optimization pipelines such as Bundle Adjustment (BA)~\cite{hartley2003multiple}, 
recent feed-forward 3D models~\cite{wang2024dust3r, tang2025mvdust3r, wang2025vggt} directly leverage end-to-end transformer architectures to learn from large-scale labeled 3D data, eschewing geometric post-processing almost entirely. However, collecting massive and high-quality 3D data remains costly and labor-intensive, posing a major limitation to the scalability of such data-driven approaches.

To enable more scalable and generalizable 3D reconstruction, self-supervised novel view synthesis (NVS) methods~\cite{huang2025spfsplat, huang2025spfsplatv2, jiang2025rayzer, hong2024pf3plat, kang2025selfsplat} jointly learn 3D scene representations and camera poses without any  3D or pose annotations. This is typically achieved by rendering novel views using self-predicted poses and enforcing photometric consistency supervision in image space.
However, in the absence of ground-truth supervision, the inherent chicken-and-egg optimization problem becomes even more challenging, often leading to instability or convergence toward degenerate solutions.
To address this, some approaches perform scene reconstruction and rendering entirely in {latent} space~\cite{jiang2025rayzer, mitchel2025true}, thereby circumventing explicit 3D optimization. Others, based on {explicit} 3D Gaussian Splatting (3DGS)~\cite{kerbl20233dgs}, rely on pretrained geometric priors~\cite{leroy2024mast3r, wang2025vggt, piccinelli2024unidepth, detone2018superpoint, lindenberger2023lightglue} obtained from supervised tasks, either for initialization~\cite{huang2025spfsplat, huang2025spfsplatv2, hong2024pf3plat} or as pseudo-labels~\cite{jiang2025anysplat}, to ensure convergence.
Moreover, despite being pose-free, most of the existing approaches~\cite{huang2025spfsplat, huang2025spfsplatv2, hong2024pf3plat, kang2025selfsplat} still require ground-truth camera intrinsics to well-condition optimization, which impedes their scalability to uncalibrated, in-the-wild imagery.
This raises a fundamental question:
\textit{Can a deep network jointly learn explicit 3D geometry and camera poses entirely from multi-view images, without any ground-truth annotations or pretrained priors?}

In this paper, we present \textbf{\ours}, a “From \underline{N}one to \underline{A}ll” \underline{S}elf-supervised \underline{3}D \underline{R}econstruction framework that jointly predicts 3D Gaussian primitives and camera parameters from uncalibrated, unposed multi-view images, requiring \textbf{no ground-truth annotations} and \textbf{no pretrained priors} during training.
\ours~builds on a standard transformer~\cite{vaswani2017transformer} architecture equipped with masked attention~\cite{huang2025spfsplatv2} to prevent target information leakage by regulating cross-view interactions. All 3D quantities, including {depth, camera intrinsics and extrinsics, and Gaussian parameters}, are inferred from a shared transformer backbone via parallel prediction heads, encouraging consistent geometric reasoning across tasks~\cite{wang2025vggt, huang2025spfsplat}. Per-view depth maps are lifted into 3D points as Gaussian centers using the estimated camera parameters. 
Unlike prior methods that directly regress unconstrained 3D points and therefore rely on pretrained priors for stable optimization~\cite{huang2025spfsplat,ye2025noposplat}, our depth-based formulation introduces stronger geometric constraints and a more well-conditioned optimization landscape. 
The entire system is trained end-to-end through photometric supervision on novel views rendered with the self-predicted camera parameters. Conceptually, \textit{the transformer backbone implicitly establishes feature correspondences through attention mechanisms, while the differentiable GS renderer performs a form of photometric BA}~\cite{lin2019photometric,delaunoy2014photometric,alismail2016photometric,lin2021barf}.

In line with the principle that well-initialized parameters and calibrated intrinsics facilitate more stable and accurate BA, we further demonstrate the flexibility and compatibility of our approach: it can be seamlessly integrated with existing state-of-the-art transformer-based 3D models~\cite{wang2025vggt, leroy2024mast3r, tang2025mvdust3r, wang2024dust3r}, and can  leverage available pretrained priors or calibrated intrinsics to achieve improved performance. As a result, \ours~offers a practical route for scaling supervised 3D models to large-scale unlabeled data.

\ours~achieves superior performance compared to other self-supervised methods and delivers results competitive with supervised approaches in novel view synthesis.
The gains in pose and depth estimation are even more substantial relative to prior methods, while
incorporating priors and camera intrinsics further enhances performance.

\section{Related Works}
\label{sec:related}
 The output of 3D reconstruction can take various forms, including point clouds~\cite{wang2024vggsfm, wang2025vggt, wang2024dust3r, leroy2024mast3r, tang2025mvdust3r, yang2025fast3r, hartley2003multiple}, depths~\cite{piccinelli2024unidepth, keetha2025mapanything, wang2025vggt, yang2024depthanything}, latent scene representations~\cite{sajjadi2022upsrt, jin2024lvsm, jiang2025rayzer, kani2024upfusion, wang2025less, mitchel2025true, sajjadi2023rust, wang2025less}, neural radiance fields (NeRFs)~\cite{chen2021mvsnerf, xu2024murf, hong2024coponerf, mildenhall2021nerf}, and explicit 3D Gaussian splats~\cite{kerbl20233dgs, ye2025noposplat, huang2025spfsplat, huang2025spfsplatv2, chen2024mvsplat, charatan2024pixelsplat}.
Based on the need for ground-truth 3D annotations or camera poses during training, these methods can be broadly categorized as supervised or self-supervised.

\subsection{Supervised 3D Reconstruction}
Supervised approaches typically incorporate 3D inductive biases into network architectures by leveraging classical geometry-based principles such as structure-from-motion (SfM)~\cite{wang2024vggsfm, brachmann2024scr, huang2023drkf}, multi-view stereo~\cite{chen2021mvsnerf, xu2024murf, chen2024mvsplat}, or epipolar geometry~\cite{charatan2024pixelsplat, he2020epipolar}.
Another line of research~\cite{wang2025vggt, wang2024dust3r, leroy2024mast3r, tang2025mvdust3r, yang2025fast3r, ye2025noposplat, sajjadi2022upsrt, kani2024upfusion, zhang2024gslrm, tang2024lgm, xu2024grm, jin2024lvsm} avoid inductive biases by employing a standard transformer~\cite{dosovitskiy2021vit} to capture cross-view relationships through attention mechanisms.
For example, DUSt3R~\cite{wang2024dust3r}, its extensions~\cite{leroy2024mast3r, tang2025mvdust3r, yang2025fast3r}, and VGGT~\cite{wang2025vggt} employ transformers for dense 3D reconstruction from unconstrained images, trained on large-scale datasets with ground-truth depth and camera poses.

Recent methods remove the need for ground-truth depth supervision by learning from photometric supervision for novel view synthesis. These models represent scenes as latent features~\cite{sajjadi2022upsrt, kani2024upfusion, jin2024lvsm}, neural volumes~\cite{jiang2024leap, wang2024pflrm}, or explicit 3D Gaussian primitives~\cite{ye2025noposplat, zhang2024gslrm, tang2024lgm, xu2024grm}.
However, accurate camera poses remain indispensable during both training and/or inference through Pl\"ucker ray embeddings~\cite{zhang2024gslrm, tang2024lgm, xu2024grm}, image rendering losses~\cite{jiang2024leap, wang2024pflrm, smart2024splatt3r, ye2025noposplat, sajjadi2022upsrt, kani2024upfusion}, pose prediction loss~\cite{hong2024coponerf}, or coarse initialization~\cite{lin2021barf, truong2023sparf}. Since these poses are typically obtained via SfM, this reliance incurs substantial computational cost and is often unreliable in sparse-view, low-overlap, or textureless scenarios. Consequently, dependence on ground-truth 3D data and camera annotations limits scalability to large-scale, in-the-wild datasets. To address this, our work focuses on fully self-supervised 3D reconstruction. 

\subsection{Self-Supervised 3D Reconstruction}
To learn from unlabeled data, self-supervised learning leverages inherent data regularities to automatically construct surrogate supervision.  In 3D vision, self-supervised monocular depth estimation~\cite{bian2019unsupervised,zhou2017unsupervised,godard2019digging} leverages photometric consistency across consecutive frames and uses view warping to provide supervisory signals without requiring ground-truth depth or camera poses.
Another major paradigm is masked content modeling in NLP and 2D vision~\cite{devlin2019bert, he2022mae, bachmann2022multimae, li2021mst, bao2021beit}, which CroCo and its variants extend to 3D using a cross-view completion task~\cite{weinzaepfel2022croco, weinzaepfel2023crocov2}.

Similarly, NVS can be viewed as a generalized masking problem in 3D space: the model reconstructs unobserved (“masked”) target views from observed context views. 
Consequently, photometric consistency on target images naturally serves as a self-supervisory signal.
Recent feed-forward approaches pursue self-supervised NVS by eliminating any ground-truth poses involved in training by reconstructing scenes from unposed images and using self-predicted poses instead of the ground-truth ones for rendering~\cite{kang2025selfsplat,hong2024pf3plat,huang2025spfsplat,huang2025spfsplatv2,jiang2025rayzer}.
However, jointly optimizing scene representations and camera motion remains a longstanding challenge. Classical structure-from-motion (SfM)~\cite{hartley2003multiple} addresses this problem through iterative optimization, alternating between local registration and global bundle adjustment (BA) of both structure and cameras.

To bypass this optimization complexity, RayZer~\cite{jiang2025rayzer} performs scene representation and rendering entirely in latent space, using predicted camera parameters to condition a geometry-free transformer-based reconstructor. While this design achieves photorealistic renderings, it does not infer
transferable camera poses, and are instead prone to interpolating context frames, as noted in~\cite{mitchel2025true}. Moreover, the lack of explicit 3D structure limits its applicability to geometry-dependent downstream tasks~\cite{yan2024gsslam,mao2025poma}. 
In contrast, methods based on explicit 3D Gaussian splatting introduce explicit geometric constraints, enabling the joint optimization of scene structure and camera poses~\cite{huang2025spfsplat,huang2025spfsplatv2,hong2024pf3plat,kang2025selfsplat,jiang2025anysplat}. For instance, PF3plat~\cite{hong2024pf3plat} and SelfSplat~\cite{kang2025selfsplat} employ separate modules for pose prediction and Gaussian reconstruction, while SPFSplat~\cite{huang2025spfsplat, huang2025spfsplatv2} unifies both within a shared backbone to enforce consistent geometric learning.
Nevertheless, most existing methods still rely on supervised pretrained priors, either for initialization~\cite{huang2025spfsplat,huang2025spfsplatv2,hong2024pf3plat} or for generating pseudo labels~\cite{jiang2025anysplat}, to ensure stable convergence.  Although SelfSplat does not require supervised priors, it still depends on self-supervised CroCoV2 initialization, and also depends on ground-truth camera intrinsics during training, similar to~\cite{huang2025spfsplat,huang2025spfsplatv2,hong2024pf3plat,kang2025selfsplat}.This dependence on camera intrinsics prevents fully self-supervised learning from uncalibrated, in-the-wild images.

Tab.~\ref{tab:method_comparison} compares training and inference requirements of representative supervised and self-supervised 3DGS methods.
Our NAS3R is a self-supervised approach that jointly learns 3D Gaussians and camera parameters without pretrained priors or ground-truth labels, while reconstructing 3D scenes from unconstrained images at inference.
\begin{table}[!t]
    \footnotesize
    \setlength{\tabcolsep}{5pt}
    \centering
    \caption{Training and inference requirements of different methods. Training columns indicate the need for ground-truth camera poses and intrinsics, while inference columns indicate whether context-view camera parameters are required as model inputs.}
    \vspace{-1em}
    \begin{tabular}{l c cc cc}
    \toprule
    \multirow{2}{*}{\textbf{Method}} &
    \multirow{2}{*}{\textbf{Priors}} &\multicolumn{2}{c}{\textbf{Training}} & \multicolumn{2}{c}{\textbf{Inference}} \\
    \cmidrule(lr){3-4} \cmidrule(lr){5-6}
    & & Pose & Intrinsics & Pose & Intrinsics \\
    \midrule
    \rowcolor{gray!20} \multicolumn{6}{l}{\textit{Supervised}} \\ 
    pixelSplat & DINO~\cite{caron2021dino} & \cmark & \cmark & \cmark & \cmark \\
    MVSplat &  UniMatch~\cite{xu2023unimatch} & \cmark & \cmark & \cmark & \cmark \\
    NoPoSplat & MASt3R~\cite{leroy2024mast3r} & \cmark & \cmark & \xmark & \cmark \\
    \rowcolor{gray!20} \multicolumn{6}{l}{\textit{Self-Supervised}} \\ 
    SelfSplat & CrocoV2~\cite{weinzaepfel2023crocov2} & \xmark & \cmark & \xmark & \cmark \\
    PF3plat & Mixed & \xmark & \cmark & \xmark & \cmark \\
    SPFSplat & MASt3R~\cite{leroy2024mast3r} & \xmark & \cmark & \xmark & \cmark \\
    SPFSplatV2 & MASt3R~\cite{leroy2024mast3r} & \xmark & \cmark & \xmark & \cmark \\
    SPFSplatV2-L & VGGT~\cite{wang2025vggt} & \xmark & \cmark & \xmark & \cmark \\
   \textbf{\ours} (Ours) &  \xmark & \xmark & \xmark & \xmark & \xmark \\
    \bottomrule
    \end{tabular}
    \label{tab:method_comparison}
    \vspace{-1.2em}
\end{table}
\section{Method}
\label{sec:method}

\begin{figure*}[t]
    \centering
    \includegraphics[width=1\textwidth]{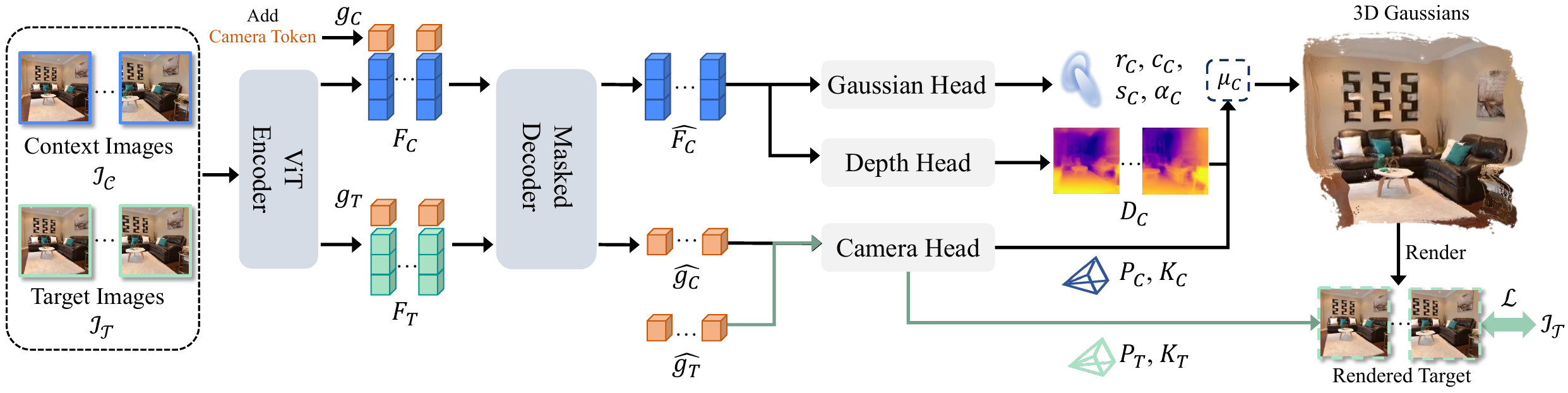}
    \vspace{-18pt}
   \caption{Training pipeline of \ours. Subscripts ``$C$'' and ``$T$'' denote context and target views, respectively. Unconstrained images are patchified into visual tokens and concatenated with a learnable camera token for camera prediction. A masked decoder regulates cross-view interactions and prevents target-to-context leakage. Refined context tokens are then processed by the Gaussian head to predict Gaussian parameters, while a depth head estimates depth maps that are lifted into 3D Gaussian centers using the predicted context poses. The predicted target poses are finally used to render novel views, providing photometric supervision for end-to-end training.}
    \label{fig:overview}
    \vspace{-15pt}
\end{figure*}
Our goal is to achieve self-supervised 3D reconstruction via novel view synthesis. To this end, we employ a feed-forward network that reconstructs 3D Gaussian primitives from unconstrained images while simultaneously predicting the corresponding camera parameters. By jointly optimizing all 3D quantities through photometric consistency on target views rendered with self-predicted camera parameters, the framework eliminates the need for any ground-truth annotations and pretrained priors.

\subsection{Problem Formulation}
Following standard training protocols for novel view synthesis~\cite{charatan2024pixelsplat,chen2024mvsplat,huang2025spfsplat}, a 3D scene is reconstructed from a set of \textit{{context}} images and subsequently used to render novel \textit{{target}} views for supervision. We denote the $V_C$ context images as $\mathcal{I_C}$ and the $V_T$ target images as $\mathcal{I_T}$. During self-supervised learning, \ours~takes both context and target images as input, denoted as $\mathcal{I}=\mathcal{I_C} \cup \mathcal{I_T}= \{\boldsymbol{I}^v\}_{v=1}^{V}$, where the total number of input images $V = V_C + V_T$.

\noindent\textbf{3D Gaussian Reconstruction.} The Gaussians are reconstructed from the \textit{{context}} images through the model $f_{\boldsymbol{\theta}}$,  expressed in the camera coordinate frame of the first view $\boldsymbol{I}^1$:
\begin{equation}
    f_{\boldsymbol{\theta}} : \mathcal{I_C} \mapsto \{   \boldsymbol{\mathcal{G}}^{v }\}_{v=1}^{V_C}, 
\label{eq:gaussian_formulation}
\end{equation}
where $\boldsymbol{\mathcal{G}}^{v}$ represents the pixel-aligned Gaussians reconstructed from each context view. 

\noindent\textbf{Camera Prediction.}  For \textit{{all}} views, a camera predictor $f_{\boldsymbol{\phi}}$ estimates both the intrinsics $\boldsymbol{K}^{v} \in \mathbb{R}^{3 \times 3}$ and the relative extrinsics $\boldsymbol{P}^{v}=[\boldsymbol{R}^{v} | \boldsymbol{T}^{v} ]$ from each view to the reference view, where $\boldsymbol{R}^{v} \in \mathbb{R}^{3 \times 3}$ represents the rotation matrix, and $\boldsymbol{T}^{v} \in \mathbb{R}^{3 \times 1}$ represents the translation vector.
\begin{equation}
    f_{\boldsymbol{\phi}} : \mathcal{I} \mapsto \{\boldsymbol{K}^v, \boldsymbol{P}^{v}\}_{v=1}^{V}.
\label{eq:pose_head}
\end{equation}

\noindent\textbf{Self-Supervised NVS.}
Novel \textit{target} views are then rendered from the reconstructed Gaussians using the predicted camera parameters of the \textit{{target}} views:
\begin{equation}
    \begin{aligned}
    \mathcal{R}: \{\boldsymbol{\mathcal{G}}^{v}\}_{v=1}^{V_C}, \{\boldsymbol{K}^v, \boldsymbol{P}^{v}\}_{v=1+V_C}^{V} \mapsto \hat{\mathcal{I_T}},
    \end{aligned}
\label{eq:image_rendering}
\end{equation}
where $\mathcal{R}$ denotes the rendering function and $\hat{\mathcal{I_T}}$ represents the rendered target views.
The model is trained by supervising the rendered images with the ground-truth target images $\mathcal{I_T}$. The training objective combines MSE and LPIPS~\cite{zhang2018lpips} losses, defined as:
\begin{equation}
    \mathcal{L}_{\text{render}} = \frac{1}{V_T}\sum\| \boldsymbol{I} - \hat{\boldsymbol{I}}\|_2 + \gamma \text{LPIPS}(\boldsymbol{I}, \hat{\boldsymbol{I}}),
\label{eq:rendering_loss}
\end{equation}
where $\boldsymbol{I} \in \mathcal{I_T}$, $\hat{\boldsymbol{I}} \in \hat{\mathcal{I_T}}$, and $\gamma$ is a weighting coefficient that balances pixel-level fidelity and perceptual similarity.

\subsection{Architecture}
Our self-supervised learning pipeline is built upon an standard transformer architecture comprising an encoder, decoder, and multiple parallel prediction heads.

\noindent\textbf{Encoder.}
For each input view $v \in [1, V]$, the corresponding image $\boldsymbol{I}^v$ is patchfied and projected into a sequence of visual tokens. These per-view tokens are processed by a shared-weight ViT~\cite{dosovitskiy2021vit} encoder to produce feature embeddings $\boldsymbol{F}^{v} \in \mathbb{R}^{L \times C}$, where $L$ and $C$ denote the number of tokens and feature dimensions, respectively. $\boldsymbol{F}^{v}$ are then concatenated with a learnable camera token $\boldsymbol{g}^v \in \mathbb{R}^{1 \times C}$, forming the decoder input $\boldsymbol{F}^v := [\boldsymbol{g}^v, \boldsymbol{F}^v]$.

\noindent\textbf{Masked Decoder.} A ViT decoder equipped with cross-attention is used to enable cross-view information exchange and aggregation.
Inspired by~\cite{huang2025spfsplatv2}, we adopt masked attention to explicitly regulate cross-view interactions and prevent unintended target information leakage. Specifically, context tokens are restricted to attend only to context tokens, ensuring that Gaussian reconstruction remains independent of target-view information. Conversely, target tokens can attend to both context and target tokens, allowing the model to leverage global scene cues for more accurate pose estimation. For each view $v$, this can be formulated as:
\begin{equation}
\boldsymbol{G}^{v} = \mathrm{MaskedDecoder}(\boldsymbol{F}^{v}, \boldsymbol{F}^{1:K}),
\label{eq:decoder}
\end{equation}
where $\boldsymbol{F}^{1:K}$ denotes the set of feature tokens from all accessible views for $\boldsymbol{F}^{v}$ in cross-attention, i.e., $K=V_C$ if view $v$ is from context views and $K=V$ for $v$ from target views. The output refined tokens $\boldsymbol{G}^{v}$ are further decomposed into refined image tokens $\hat{\boldsymbol{F}^{v}}$ and a refined camera token $\hat{\boldsymbol{g}^{v}}$.

\noindent\textbf{Camera Predictor.} The camera head predicts the per-view camera extrinsics $\boldsymbol{P}^{v}$ and intrinsics $\boldsymbol{K}^{v}$ from the refined camera tokens $\hat{\boldsymbol{g}^{v}}$.
The predicted poses are normalized such that the first input view is assigned as the canonical reference frame.
Camera intrinsics are parameterized by the field of view (FOV), assuming identical FOV along the $x$ and $y$ axes and a principal point located at the image center.
 
\noindent\textbf{Gaussian Predictor.}
To regress the center $\boldsymbol{\mu} \in \mathbb{R}^3$ for each Gaussian primitive, existing methods typically adopt either a canonical-space paradigm~\cite{ye2025noposplat,huang2025spfsplat,huang2025spfsplatv2}, which directly predicts canonical 3D points as Gaussian centers, or a local-to-global paradigm~\cite{chen2024mvsplat,charatan2024pixelsplat,kang2025selfsplat,hong2024pf3plat}, which lifts per-view depth predictions into global 3D coordinates.
However, canonical-space formulations~\cite{ye2025noposplat,huang2025spfsplat} struggle to learn canonical 3D point locations without explicit 3D point initialization or supervision. This is because canonical 3D points implicitly encode scene geometry, camera poses, and camera intrinsics, yet these quantities are not explicitly constrained during training, leading to an ill-conditioned optimization problem. Consequently, these methods rely on either initialization from pretrained priors or distillation from pretrained DUSt3R~\cite{wang2024dust3r} point predictions to provide a warm-up phase during early training when training from random initialization.

Therefore, to enable training without pretrained priors, we adopt a local-to-global paradigm that derives 3D Gaussian centers from normalized camera poses, valid camera intrinsics, and positive depth values. This results in a physically grounded and well-conditioned formulation that promotes stable and consistent convergence. Moreover, by construction, the Gaussian centers are constrained to lie along the corresponding viewing rays, providing stronger geometric guidance than direct canonical-point regression.
Specifically, we employ a DPT-based~\cite{ranftl2021dpt} depth head to predict per-pixel depths from the refined context image tokens $\hat{\boldsymbol{F}}^{v}$. The predicted depths are passed through a sigmoid function, interpolated between near and far planes, and subsequently lifted into 3D space using the predicted poses $\boldsymbol{P}^{v}$ and intrinsics $\boldsymbol{K}^{v}$ to obtain the Gaussian centers.

The Gaussian parameter head follows a similar DPT-style architecture. It takes as input both the refined context image tokens and the original context images to preserve fine-grained spatial details, and predicts each Gaussian’s rotation quaternion $\boldsymbol{r} \in \mathbb{R}^4$, scale $\boldsymbol{s} \in \mathbb{R}^3$, opacity $\alpha \in \mathbb{R}$, and spherical harmonics (SH) coefficients $\boldsymbol{c} \in \mathbb{R}^k$, where $k$ denotes the number of degrees of freedom.


\noindent\textbf{Model Compatibility.}
Since \ours{} shares a similar architecture with recent reconstruction models, including 3R-based frameworks~\cite{wang2024dust3r,leroy2024mast3r,tang2025mvdust3r} and VGGT~\cite{wang2025vggt}, we adopt their designs to ensure compatibility. Specifically, we equip their decoders with masked attention to regulate cross-view information flow.
For VGGT, we add a Gaussian parameter head while retaining all other components. For 3R-based models, we build upon the multi-view extension of MASt3R~\cite{leroy2024mast3r}, following~\cite{huang2025spfsplat,huang2025spfsplatv2,ye2025noposplat}, and add depth, Gaussian parameter, and camera prediction heads. The camera head is implemented as a lightweight three-layer MLP that predicts rotation using a 6D representation~\cite{zhou2019continuity} and translation using four homogeneous coordinates.

Since camera intrinsics are often available from sensor metadata, we also study a variant with known intrinsics in Sec.~\ref{sec:results_with_priors} to reduce scale ambiguity. In this variant, intrinsics are embedded through a linear layer and concatenated with the camera and encoder tokens before decoding.

\subsection{Training and Inference}
During training, the entire pipeline is optimized end-to-end using the photometric consistency loss defined in Eq.~\ref{eq:rendering_loss}.
The optimization  is enabled by CUDA-based 3D Gaussian Splatting renderer that supports gradient computation with respect to camera poses, intrinsics, and Gaussian parameters. Conceptually, the renderer performs a form of photometric bundle adjustment, where each Gaussian primitive must explain consistent image observations across multiple views. Consequently, rendering errors are propagated back to both the scene representation and camera parameters, driving the model to jointly infer geometrically consistent cameras, and reconstruct a coherent 3D scene that best explains the observed images.

 At inference time, the model requires only the context images as input to predict their corresponding camera parameters and reconstruct the scene.

\section{Experiments}
\label{sec:experiments}


\subsection{Experimental Settings}
\label{sec:experimental_settings}
\textbf{Dataset.} We train our method on RealEstate10K (RE10K)~\cite{zhou2018stereo} and DL3DV~\cite{ling2024dl3dv}. To evaluate generalization across diverse scenes, we additionally test on the object-centric DTU dataset~\cite{jensen2014dtu}, the outdoor ACID dataset~\cite{liu2021infinite}, and ScanNet++~\cite{yeshwanth2023scannetpp}. For downstream finetuning, we further train and evaluate on BlendedMVS~\cite{yao2020blendedmvs}.

\noindent\textbf{Baselines.}
For NVS methods, we first compare against approaches that do not rely on supervised 3D reconstruction priors (Sec.~\ref{sec:results_no_priors}). Although the original NoPoSplat and SPFSplat are typically initialized from the supervised MASt3R model, their papers show that a brief warm-up stage using a DUSt3R-based point cloud distillation loss enables training from random initialization. Since this distillation is used only during early training, we denote the resulting variants with a superscript $^\ast$. SelfSplat employs self-supervised CroCoV2\cite{weinzaepfel2023crocov2} features, pixelSplat adopts DINO~\cite{caron2021dino} features, and MVSplat uses the supervised UniMatch~\cite{xu2023unimatch} feature extractor. However, these methods do not leverage supervised 3D predictions (e.g., depth or pointmaps) for initialization or supervision. Accordingly, we compare against pixelSplat~\cite{charatan2024pixelsplat}, MVSplat~\cite{chen2024mvsplat}, NoPoSplat$^\ast$\cite{ye2025noposplat}, SelfSplat\cite{kang2025selfsplat}, and SPFSplat$^\ast$~\cite{huang2025spfsplat}.
In Sec.~\ref{sec:results_with_priors}, we further compare against methods that leverage supervised 3D reconstruction priors, including the original MASt3R-initialized NoPoSplat and SPFSplat, as well as PF3plat~\cite{hong2024pf3plat} and SPFSplatV2/V2-L~\cite{huang2025spfsplatv2}.

For camera pose estimation in Sec.~\ref{sec:results_no_priors} and Sec.~\ref{sec:results_with_priors}, we compare against (1) SfM-based approaches, including SuperPoint~\cite{detone2018superpoint} + SuperGlue~\cite{sarlin2020superglue}, DUSt3R~\cite{wang2024dust3r}, MASt3R~\cite{leroy2024mast3r}, and VGGT~\cite{wang2025vggt}; and (2) NVS-based approaches, including NoPoSplat, SelfSplat, PF3plat, SPFSplat, and SPFSplatV2/V2-L. For depth estimation, we compare against MVSplat, NoPoSplat$^\ast$ and SPFSplat$^\ast$.



\definecolor{lightgray}{gray}{0.5}
\begin{table*}[!ht]
    \centering
    \caption{Performance comparison of in-domain and out-of-domain two-view novel view synthesis for models trained with \textbf{no supervised 3D priors}. The \textbf{best} results among self-supervised methods are highlighted. $\ast$ indicates the variants with random initialization. }
    \vspace{-1em}
    \resizebox{\textwidth}{!}{%
    \begin{tabular}{l c ccc ccc ccc ccc}
    \toprule
    \multirow{2}{*}{\textbf{Method}} &  \multirow{2}{*}{\textbf{Training Data}} &
    \multicolumn{3}{c}{\textbf{RE10K}} &
    \multicolumn{3}{c}{\textbf{ACID}} &
    \multicolumn{3}{c}{\textbf{DL3DV}} &
    \multicolumn{3}{c}{\textbf{DTU}} \\
    \cmidrule(lr){3-5} \cmidrule(lr){6-8} \cmidrule(lr){9-11} \cmidrule(lr){12-14}
    && PSNR $\uparrow$ & SSIM $\uparrow$ & LPIPS $\downarrow$
    & PSNR $\uparrow$ & SSIM $\uparrow$ & LPIPS $\downarrow$
    & PSNR $\uparrow$ & SSIM $\uparrow$ & LPIPS $\downarrow$
    & PSNR $\uparrow$ & SSIM $\uparrow$ & LPIPS $\downarrow$ \\
    \midrule
    \rowcolor{gray!20} \multicolumn{14}{l}{\textit{Supervised}} \\ 
\textcolor{lightgray}{pixelSplat} & \multirow{3}{*}{RE10K} &
    \textcolor{lightgray}{23.859} & \textcolor{lightgray}{0.808} & \textcolor{lightgray}{0.184} &
    \textcolor{lightgray}{25.477} & \textcolor{lightgray}{0.770} & \textcolor{lightgray}{0.207} &
    \textcolor{lightgray}{21.370} & \textcolor{lightgray}{0.713} & \textcolor{lightgray}{0.250} &
    \textcolor{lightgray}{15.067} & \textcolor{lightgray}{0.539} & \textcolor{lightgray}{0.341} 
    \\
\textcolor{lightgray}{MVSplat} & &
    \textcolor{lightgray}{24.012} & \textcolor{lightgray}{0.812} & \textcolor{lightgray}{0.175} &
    \textcolor{lightgray}{25.525} & \textcolor{lightgray}{0.773} & \textcolor{lightgray}{0.199} &
    \textcolor{lightgray}{20.344} & \textcolor{lightgray}{0.673} & \textcolor{lightgray}{0.258} &
    \textcolor{lightgray}{14.542} & \textcolor{lightgray}{0.537} & \textcolor{lightgray}{0.324} 
    \\
\textcolor{lightgray}{NoPoSplat$^{\ast}$} & &
    \textcolor{lightgray}{21.382} & \textcolor{lightgray}{0.709} & \textcolor{lightgray}{0.266} &
    \textcolor{lightgray}{23.193} & \textcolor{lightgray}{0.667} & \textcolor{lightgray}{0.274} &
    \textcolor{lightgray}{18.235} & \textcolor{lightgray}{0.518} & \textcolor{lightgray}{0.364} &
    \textcolor{lightgray}{15.113} & \textcolor{lightgray}{0.455} & \textcolor{lightgray}{0.450} \\
    \rowcolor{gray!20} \multicolumn{14}{l}{\textit{Self-Supervised}} \\ 
    SelfSplat & \multirow{3}{*}{RE10K} & 19.152 & 0.680 & 0.328 & 22.204 & 0.686 & 0.316 & 17.159 & 0.538 & 0.400  & 13.249 & 0.434 & 0.441 \\
    SPFSplat$^{\ast}$ & & 21.306 & 0.693 & 0.248 & 23.354 & 0.662 & 0.260 & 18.091	& 0.480	& 0.358 & 14.042 & 0.432 & 0.426 \\
    \textbf{\ours} (Ours) & & \textbf{23.130}	& \textbf{0.764}	& \textbf{0.193}	& \textbf{25.030}	& \textbf{0.734}	& \textbf{0.209}	& \textbf{19.507} &	\textbf{0.573}	& \textbf{0.293}	& \textbf{15.229} &	\textbf{0.524}	& \textbf{0.317} \\
    \midrule
    \rowcolor{gray!20} \multicolumn{14}{l}{\textit{Self-Supervised}} \\ 
    SelfSplat & \multirow{2}{*}{DL3DV} & 19.371	& 0.682	& 0.379	& 22.410 &	0.678	& 0.348	& 18.685	& 0.585	
    & 0.417 &	12.765 &	0.442	& 0.500 \\
    \textbf{\ours} (Ours) & & \textbf{21.351}	& \textbf{0.693} &	\textbf{0.245}	& \textbf{24.178} &	\textbf{0.697} & \textbf{0.229}	& \textbf{20.069}	& \textbf{0.588}	& \textbf{0.281} &	\textbf{15.511}	& \textbf{0.519}	 & \textbf{0.319}\\
    \bottomrule
    \end{tabular}%
    }
    \label{tab:rek_results}
\end{table*}

\noindent\textbf{Evaluation Protocol.}
For novel view synthesis, we report pixel-level PSNR, patch-level SSIM~\cite{wang2004ssim}, and feature-level LPIPS~\cite{zhang2018lpips}. Unless otherwise specified, target images are rendered using predicted poses, jointly evaluating Gaussian reconstruction and pose estimation. For pose estimation, following prior work~\cite{sarlin2020superglue,ye2025noposplat}, we report the area under the cumulative pose error curve (AUC) at different thresholds, where the pose error is defined as the maximum of the rotation and translation angular errors. For depth estimation, we report Absolute Relative Error (rel) and Inlier Ratio ($\tau$) at a threshold of 1.25~\cite{eigen2014depth,uhrig2017sparsity}.

\begin{figure}[t]
    \centering
    \includegraphics[width=\columnwidth]{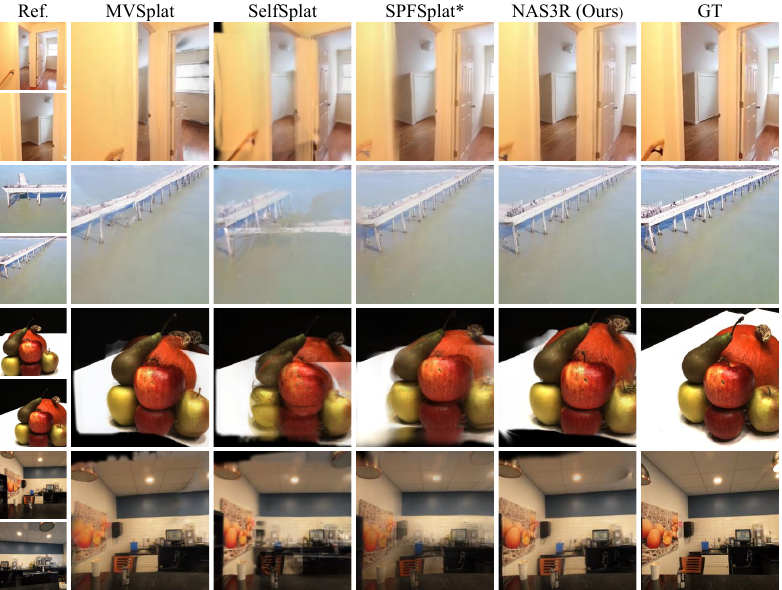}
    \vspace{-15pt}
    \caption{Comparison of NVS results across different methods. The leftmost column shows the two-view context images. From top to bottom, the settings are RE10K, RE10K$\rightarrow$ACID, RE10K$\rightarrow$DTU, and RE10K$\rightarrow$DL3DV.}
    \label{fig:nvs_vis}
    \vspace{-15pt}
\end{figure}

\subsection{Implementation Details}\vspace{-0.4em}
\ours~is implemented in PyTorch and 
all models are trained on a single NVIDIA A100 GPU. Each training sample corresponds to a scene comprising context and target views, where the frame interval between context frames is progressively increased during training. To ensure stable convergence, the camera prediction head is initialized to output identity poses and a focal length equal to the input image size. We assume that all views within a scene share the same camera intrinsics, with identical field of view (FOV) along the $x$ and $y$ axes. Training is performed at resolutions of $224 \times 224$ and $256 \times 256$ for the VGGT-based and MASt3R-based variants, respectively. Unless stated otherwise, we use a VGGT-based architecture as the default backbone, with the default input consisting of two views.

\begin{figure}[!th]
    \centering
    \includegraphics[width=\columnwidth]{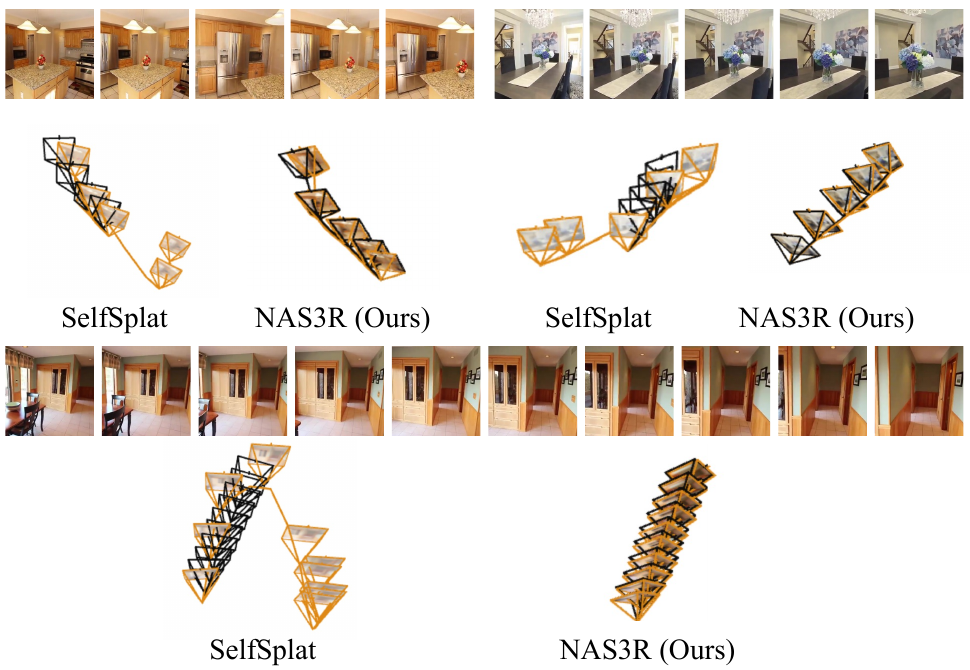}
    \vspace{-15pt}
    \caption{Visual comparison of pose trajectories on RE10K. Camera frustums for ground-truth and predicted poses are shown in black and \textcolor{customorange}{orange}, respectively. The top two examples correspond to 5-view reconstruction, while the bottom example corresponds to 10-view reconstruction.}
    \label{fig:pose_traj}
    \vspace{-15pt}
\end{figure}

\begin{table*}[!ht]
\footnotesize
\centering
\caption{Performance comparison of two-view pose estimation (AUC, \%) on RE10K, ACID and DL3DV datasets. All NVS models are trained on RE10K with \textbf{no supervised 3D priors}. $\ast$ indicates the SPFSplat variant trained with random initialization.}
\setlength{\tabcolsep}{3.2pt}
\vspace{-1em}
\begin{tabular}{l cc cc cc cc cc cc cc cc cc}
    \toprule
    \multirow{3}{*}{\textbf{Method}} 
    & \multicolumn{6}{c}{\textbf{RE10K}} 
    & \multicolumn{6}{c}{\textbf{ACID}} 
     & \multicolumn{6}{c}{\textbf{DL3DV}} \\
    \cmidrule(lr){2-7} \cmidrule(lr){8-13}
    \cmidrule(lr){14-19}
    & \multicolumn{2}{c}{Rotation} 
    & \multicolumn{2}{c}{Translation} 
    & \multicolumn{2}{c}{Overall} 
    & \multicolumn{2}{c}{Rotation} 
    & \multicolumn{2}{c}{Translation}  
    & \multicolumn{2}{c}{Overall} 
    & \multicolumn{2}{c}{Rotation} 
    & \multicolumn{2}{c}{Translation}  
    & \multicolumn{2}{c}{Overall} \\
    \cmidrule(lr){2-3} \cmidrule(lr){4-5} \cmidrule(lr){6-7}
    \cmidrule(lr){8-9} \cmidrule(lr){10-11} \cmidrule(lr){12-13} \cmidrule(lr){14-15} \cmidrule(lr){16-17} \cmidrule(lr){18-19}
    & 10$^\circ$ $\uparrow$ & 20$^\circ$ $\uparrow$ 
    & 10$^\circ$ $\uparrow$ & 20$^\circ$ $\uparrow$ 
    & 10$^\circ$ $\uparrow$ & 20$^\circ$ $\uparrow$ 
    & 10$^\circ$ $\uparrow$ & 20$^\circ$ $\uparrow$ 
    & 10$^\circ$ $\uparrow$ & 20$^\circ$ $\uparrow$  
    & 10$^\circ$ $\uparrow$ & 20$^\circ$ $\uparrow$ 
    & 10$^\circ$ $\uparrow$ & 20$^\circ$ $\uparrow$ 
    & 10$^\circ$ $\uparrow$ & 20$^\circ$ $\uparrow$  
    & 10$^\circ$ $\uparrow$ & 20$^\circ$ $\uparrow$  \\
    \midrule
     \rowcolor{gray!20} \multicolumn{19}{l}{\textit{SfM-based}} \\ 
    \textcolor{lightgray}{SP + SG} & \textcolor{lightgray}{62.7} & \textcolor{lightgray}{73.3}  & \textcolor{lightgray}{42.0} & \textcolor{lightgray}{58.5} & \textcolor{lightgray}{40.6} & \textcolor{lightgray}{56.9} & \textcolor{lightgray}{62.3} & \textcolor{lightgray}{72.4}  & \textcolor{lightgray}{38.0} & \textcolor{lightgray}{51.6} & \textcolor{lightgray}{36.3} & \textcolor{lightgray}{50.0} &
    \textcolor{lightgray}{47.8} & \textcolor{lightgray}{56.1}  & \textcolor{lightgray}{40.2} & \textcolor{lightgray}{52.6} & \textcolor{lightgray}{37.2} & \textcolor{lightgray}{49.2} \\
     \rowcolor{gray!20} \multicolumn{19}{l}{\textit{NVS-based}} \\ 
    SelfSplat & 48.7 & 62.7 & 19.2 & 32.8 & 18.4 & 31.8 & 64.5 & 75.2 & 26.4 & 40.5 & 25.8 & 39.6 & 21.0 & 34.4 & 8.9 &22.4 & 6.1 & 16.8 \\
    SPFSplat$^\ast$  & 54.9 & 67.7  & 24.8 & 40.9  & 23.9 & 39.8 & 50.6 & 65.3 &  19.5 & 36.3 & 18.2 & 34.4 & 16.6 & 32.7 & 8.8 & 12.2 & 7.1 & 15.2\\
    \textbf{\ours} & \textbf{69.9} & \textbf{78.4} & \textbf{52.7} & \textbf{66.3}  & \textbf{51.0} & \textbf{64.9}  & \textbf{66.0} & \textbf{76.6}  & \textbf{37.9} & \textbf{51.9} & \textbf{36.4} & \textbf{50.1} & \textbf{38.5} & \textbf{51.8} & \textbf{25.1} & \textbf{42.1}  & \textbf{20.5} & \textbf{34.2} \\
    \bottomrule
\end{tabular}
\label{tab:pose_estimation}
\end{table*}
        
        

\begin{figure*}[t]
\centering
\begin{minipage}[t]{0.32\textwidth}
    \centering
    \footnotesize
    \setlength{\tabcolsep}{11pt}
    \captionof{table}{Comparison of two-view depth estimation results on BlendedMVS dataset.}
    \vspace{-1em} 
    \begin{tabular}{lcc}
        \toprule
        Method & rel $\downarrow$  & $\tau$ $\uparrow$ \\
        \midrule
        \rowcolor{gray!20} \multicolumn{3}{l}{\textit{Supervised}} \\
        MVSplat & 0.405 & 54.0 \\
        NoPoSplat & 0.508 & 34.1 \\
        \rowcolor{gray!20} \multicolumn{3}{l}{\textit{Self-Supervised}} \\
        SPFSplat$^\ast$ &  0.255 & 60.3 \\
        \textbf{NAS3R} (Ours) & \textbf{0.206}	 &  \textbf{71.4} \\
        
        \bottomrule
    \end{tabular}
    \label{tab:depth_results}
\end{minipage}
\hfill
\begin{minipage}[t]{0.62\textwidth}
    \centering
    \vspace{0mm} 
    \includegraphics[width=\textwidth]{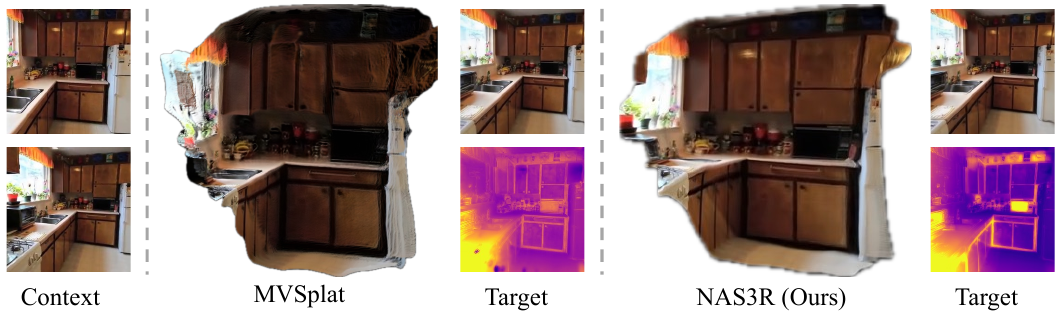}
    \vspace{-2em}
    
    \captionof{figure}{Examples of 3D Gaussians and rendered RGB and depth results on RE10K.}
    \label{fig:gs_vis}
    
\end{minipage}
\vspace{-1em}
\end{figure*}

\subsection{Results}

\label{sec:results_no_priors}
In this section, we compare NAS3R with other NVS approaches trained with \textbf{no supervised 3D priors}. Baseline settings for all methods are detailed in Sec.~\ref{sec:experimental_settings}.

\noindent\textbf{Novel View Synthesis.} 
We present in-domain and out-of-domain quantitative NVS results in Tab.~\ref{tab:rek_results}. Our approach, despite not using any ground-truth poses or intrinsics during training, still outperforms other self-supervised methods, and is comparable and in some cases superior   to supervised methods. Notably, it provides higher quality output even under wide viewpoint changes, as illustrated in Fig.~\ref{fig:nvs_vis}.

\noindent\textbf{Relative Pose Estimation.}
We evaluate 2-view relative pose estimation on the RE10K, ACID, and DL3DV datasets, as shown in Tab.~\ref{tab:pose_estimation}. To assess generalization, all NVS methods are trained on RE10K and evaluated in a zero-shot manner on the other datasets. For all NVS-based methods, predicted poses are obtained from their respective pose heads.
Despite being trained without intrinsics information, our method consistently outperforms recent self-supervised approaches, particularly on the more challenging DL3DV dataset, demonstrating its ability to learn camera poses purely from image-level signals. Remarkably, NAS3R approaches or even surpasses SuperPoint + SuperGlue, which is trained in a fully supervised manner for feature matching, which might indicate s strong capability in establishing feature correspondences in a self-supervised setting.
We also provide a qualitative comparison with SelfSplat in Fig.~\ref{fig:pose_traj}, showing more accurate pose trajectories under both 5-view and 10-view settings.

\noindent\textbf{Depth Estimation.}
As shown in Tab.~\ref{tab:depth_results}, we evaluate the quality of the predicted depth maps on BlendedMVS using RE10K-trained models. Our method achieves the best performance, even surpassing the supervised NVS method MVSplat, which explicitly leverages ground-truth camera parameters and incorporates cost volumes into its backbone. This result highlights the strong capability of a standard transformer, without any built-in 3D inductive bias, to learn accurate geometry purely from image-level supervision.

\noindent\textbf{Geometry Reconstruction.}
As shown in Fig.~\ref{fig:gs_vis}, MVSplat produces irregular and noisy Gaussian geometry, leading to smeared surfaces and inconsistent depth.
In contrast, our NAS3R reconstruction yields significantly more coherent Gaussian distributions with cleaner spatial organization. This improved structure translates into sharper, more accurate depth renderings that better preserve planar surfaces and fine details, highlighting the ability of our approach to generate stable geometry and depth.

\subsection{Leveraging Pretrained Priors}
\label{sec:results_with_priors}
Our self-supervised learning paradigm can be seamlessly applied to state-of-the-art 3D models and incorporate their pretrained priors through weight initialization. Following previous works~\cite{ye2025noposplat,huang2025spfsplat,huang2025spfsplatv2,wang2025vggt}, we integrate two representative 3D priors, MASt3R and VGGT. For MASt3R initialization, the encoder and decoder are initialized with pretrained MASt3R weights, while the depth head is initialized using the $z$-dimension of the pointmap head. For VGGT initialization, all components are intialized from the pretrained weights except for the Gaussian parameter head. 


\begin{table}[!t]
    \centering
    \footnotesize
    \caption{Performance comparison of two-view NVS when \textbf{using supervised priors}. \textbf{Best} and \underline{second-best} results are highlighted.}
    \setlength{\tabcolsep}{1.2pt}
    \vspace{-1em}
    \begin{tabular}{l c ccc ccc}
    \toprule
    \multirow{2}{*}{\textbf{Method}} 
    & \multicolumn{3}{c}{\textbf{RE10K}} 
    & \multicolumn{3}{c}{\textbf{RE10K $\rightarrow$ACID}} 
    \\
    \cmidrule(lr){2-4} \cmidrule(lr){5-7}
     & 
     PSNR $\uparrow$ & SSIM $\uparrow$ & LPIPS $\downarrow$ 
     & PSNR $\uparrow$ & SSIM $\uparrow$ & LPIPS $\downarrow$ 
      \\
    \midrule
     \rowcolor{gray!20} \multicolumn{7}{l}{\textit{Supervised}} \\ 
    NoPoSplat &  25.061 & 0.839 &0.161 & 25.765&	0.776 & 0.199  \\
     \rowcolor{gray!20} \multicolumn{7}{l}{\textit{Self-Supervised}} \\ 
    PF3plat &   21.042 & 0.739 & 0.233  & 20.726	& 0.610 & 0.308\\
    SPFSplat &   25.484	& 0.847	& 0.153	& 25.965	& 0.781 & 0.190 \\
    SPFSplatV2 &   25.693 &	0.853 &	0.149 & 26.220	& 0.789 &0.185\\
    SPFSplatV2-L &   25.668 & 0.855 & \underline{0.137} & 26.361	& 0.796 & \underline{0.169} \\
     \textbf{\ours (MASt3R)} &  \underline{25.814} &	\underline{0.856}	& 0.149 & \underline{26.492}	& \underline{0.800} & {0.183} \\
    \textbf{\ours (VGGT)} &  \textbf{25.888}	& \textbf{0.861}	& \textbf{0.136} & \textbf{26.663}	& \textbf{0.807} & \textbf{0.166} \\
    \bottomrule
    \end{tabular}
    \vspace{-7pt}
    \label{tab:rek_results_with_priors}
\end{table}

As shown in Tab.~\ref{tab:rek_results_with_priors}, incorporating pretrained priors further boosts NVS performance, allowing our method to outperform competing approaches despite using only self-supervised photometric supervision. Similarly, Tab.~\ref{tab:pose_estimation_with_priors} shows notable gains in pose estimation, where \ours{} consistently matches or surpasses other NVS-based methods and even exceeds the performance of MASt3R and VGGT themselves. When ground-truth intrinsics are additionally provided (denoted as \ours-I), performance improves further, indicating that explicit calibration information helps reduce scale ambiguity and facilitates optimization.

Comparing these results with Tab.\ref{tab:rek_results} and Tab.\ref{tab:pose_estimation}, the consistent gains in both NVS and pose estimation confirm the value of pretrained priors as a strong geometric initialization, guiding photometric bundle adjustment toward better geometric solutions. More importantly, the improvements over the pretrained models used for initialization, particularly in pose estimation, highlight the potential of self-supervised photometric optimization to further adapt supervised 3D priors to large-scale unlabeled data.

\begin{table}[!t]
\footnotesize
\centering
\setlength{\tabcolsep}{4.3pt}
   \caption{Performance comparison of two-view pose estimation when \textbf{using supervised priors}. ``-I'' indicates using GT intrinsics.}
   \vspace{-1em}
\begin{tabular}{lccccccccccc}
    \toprule
    \multirow{2}{*}{\textbf{Method}} & \multicolumn{3}{c}{\textbf{RE10K}} & \multicolumn{3}{c}{\textbf{RE10K $\rightarrow$ACID}}  \\
    \cmidrule(lr){2-4} \cmidrule(lr){5-7}  
    
    &  5$^\circ$ $\uparrow$ & 10$^\circ$ $\uparrow$ & 20$^\circ$ $\uparrow$ &  5$^\circ$ $\uparrow$ &  10$^\circ$ $\uparrow$ &  20$^\circ$ $\uparrow$  \\
    \midrule
    \rowcolor{gray!20} \multicolumn{7}{l}{\textit{SfM-based}} \\ 
    DUSt3R & 33.6 & 54.1 & 70.2 & 11.8 & 27.9 & 47.0 \\ 
    MASt3R & 28.1 & 49.4 & 67.1 & 13.8 & 31.2 & 50.7 \\
    VGGT & 25.7 & 47.4 & 65.8 & 14.2 & 30.4 & 48.6 \\
    \rowcolor{gray!20} \multicolumn{7}{l}{\textit{NVS-based}} \\ 
    NoPoSplat & 57.2 & 72.8 & 83.3 & 33.5 & 49.7 & 64.5 \\
    PF3plat  & 18.7 & 39.8 & 61.3 & 6.0 & 16.5 & 34.0 \\
    SPFSplat & 61.7 & 75.5 & 84.5 & 36.4 & 52.0 & 66.2 \\
    SPFSplatV2 & 63.8 & 77.6 & 86.3 & 38.7 & \underline{54.1} & 67.2 \\
    SPFSplatV2-L & 64.5 & \underline{78.0} & 86.4 & 37.9 & 53.9 & 67.1 \\
    \textbf{\ours (MASt3R)} & 63.8 & 77.6	& 86.4 & 35.8	& 53.3 & 67.7 \\
    \textbf{\ours (VGGT)} & 62.4 & 77.2 & 86.6 & 33.6 & 52.2	& 67.1 \\
   \textbf{\ours-I (MASt3R)} & \textbf{68.3} &	\textbf{80.3}	& \underline{87.8} & \underline{43.8} & \textbf{58.4}	& \textbf{70.7} \\
    \textbf{\ours-I (VGGT)} & \underline{67.7} &	\textbf{80.3}	& \textbf{88.1} & \textbf{44.0}	& \textbf{58.4}	& \underline{70.4} \\
    \bottomrule
    \end{tabular}
    \vspace{-7pt}
    \label{tab:pose_estimation_with_priors}
\end{table}

\begin{table}[t]
\footnotesize
\setlength{\tabcolsep}{2.4pt}
    \centering
     	\caption{Performance on NVS and pose estimation (ScanNet++) and depth estimation (BlendedMVS) with increasing training data.}
     \vspace{-1em}
    \begin{tabular}{l ccc cc cc }
        \toprule
        \multirow{2}{*}{\textbf{Training Data}} & \multicolumn{3}{c}{\textbf{NVS}} & \multicolumn{2}{c}{\textbf{Pose}} & \multicolumn{2}{c}{\textbf{Depth}}\\
        \cmidrule(lr){2-4} \cmidrule(lr){5-6} \cmidrule(lr){7-8}
        & PSNR$\uparrow$ & SSIM$\uparrow$ & LPIPS$\downarrow$ & 10$^\circ$ $\uparrow$ & 20$^\circ$ $\uparrow$ & rel $\downarrow$  & $\tau$ $\uparrow$ \\ 
        \midrule
        RE10K & 15.146 & 0.520 & 0.442 & 19.2 & 34.1 & 0.206 & 71.4 \\
        DL3DV  &  16.171  & 0.542 & 0.431 & 19.1 & 33.2 & 0.175 & 71.9 \\
        RE10K+DL3DV  & \textbf{16.316} & \textbf{0.552} & \textbf{0.420}  & \textbf{24.5} & \textbf{41.6} & \textbf{0.145} & \textbf{79.1} \\
        \bottomrule
    \end{tabular}
    \vspace{-1em}
\label{tab:scaling_data}
\end{table}

\subsection{Ablation Analysis}

\noindent\textbf{Scaling with Larger Training Sets.}
Because NAS3R does not require ground-truth labels, it naturally scales to large, unconstrained datasets. To study the effect of data scale, we train on the combined RE10K+DL3DV dataset and evaluate NVS and pose estimation on ScanNet++, as well as depth estimation on BlendedMVS. As shown in Tab.~\ref{tab:scaling_data}, increasing the amount of training data consistently improves performance across all three tasks. These results demonstrate the strong scalability of our self-supervised framework and its potential to leverage large-scale unlabeled datasets.

\noindent\textbf{More Input Views.}
As shown in Tab.~\ref{tab:multi_view_for_nvs}, quantitative results on RE10K indicate that both NVS and pose estimation performance improve consistently as the number of context views increases. This demonstrates that our method effectively leverages additional multi-view information to produce more reliable renderings and more accurate camera pose predictions.

\noindent\textbf{Self-supervised Pretraining.}
To evaluate the effectiveness of our model as a self-supervised pretraining strategy for downstream finetuning with ground-truth supervision, we compare three settings on the BlendedMVS dataset: (1) NAS3R's self-supervised training on RE10K and DL3DV using only photometric supervision, evaluated in a zero-shot setting on BlendedMVS; (2) supervised training on BlendedMVS from scratch using ground-truth depth and pose losses (following MapAnything~\cite{keetha2025mapanything}); and (3) supervised training on BlendedMVS with the same losses, but initialized from the RE10K+DL3DV pretrained weights in (1). As shown in Tab.~\ref{tab:gt_supervised_training}, NAS3R trained solely on RE10K+DL3DV achieves depth performance slightly better than the same architecture trained from scratch with ground-truth supervision, and significantly surpasses it in pose estimation. Furthermore, initializing from the self-supervised RE10K+DL3DV weights significantly outperforms random initialization when fine-tuning with supervised depth and pose losses. These results highlight the model’s ability to learn meaningful geometry purely from self-supervision and demonstrate that large-scale self-supervised pretraining provides a strong and broadly generalizable initialization for downstream supervised tasks.

\begin{table}[t]
\footnotesize
\setlength{\tabcolsep}{4pt}
    \centering
     \caption{Performance on novel view synthesis and pose estimation with varying number of input views.}
     \vspace{-1em}
    \begin{tabular}{l ccc ccc}
        \toprule
        \multirow{2}{*}{\textbf{Num of Views}} & \multicolumn{3}{c}{\textbf{NVS}} & \multicolumn{3}{c}{\textbf{Pose}} \\
        \cmidrule(lr){2-4} \cmidrule(lr){5-7}
        & PSNR$\uparrow$ & SSIM$\uparrow$ & LPIPS$\downarrow$ &  5$^\circ$ $\uparrow$ & 10$^\circ$ $\uparrow$ & 20$^\circ$ $\uparrow$  \\ 
        \midrule
        2 views & 23.130	& 0.764	& 0.193 & 32.6	& 51.0 & 64.9\\
        3 views  &  25.384  & 0.829 & 0.141 & 37.9 & 57.6 & 72.0\\
        5 views  & 25.983 & 0.844 & 0.131 & 38.9 & 58.5 & 72.8\\
        10 views  &  \textbf{27.093} &  \textbf{0.872} & \textbf{0.113} & \textbf{41.0} & \textbf{61.0} & \textbf{75.5} \\
        \bottomrule
    \end{tabular}
\label{tab:multi_view_for_nvs}
    \vspace{-0.5em}
\end{table}
\begin{table}[!t]
\footnotesize
\centering
\setlength{\tabcolsep}{3pt}
  \caption{Self-supervised pretraining improves downstream finetuning. Supervised finetuning from self-supervised pretrained weights (3) outperforms both self-supervised learning (1) and supervised training from scratch (2) on BlendedMVS.}
   \vspace{-1em}
    \begin{tabular}{l cc ccc }
    \toprule
    \multirow{2}{*}{\textbf{Method}} & \multicolumn{2}{c}{\textbf{Depth}} & \multicolumn{3}{c}{\textbf{Pose}}  \\
    \cmidrule(lr){2-3} \cmidrule(lr){4-6}  
    
    &  rel $\downarrow$  & $\tau$ $\uparrow$ & 5$^\circ$ $\uparrow$ &  10$^\circ$ $\uparrow$ &  20$^\circ$ $\uparrow$  \\
    \midrule
    (1) Self-Supervised (NAS3R)  & 0.145 & 79.1 &	29.8	& 49.5 & 66.8 \\
    (2) Supervised from scratch  & 0.177	& 76.6	& 10.7&	23.5	& 39.4 \\
    (3) Supervised from weights in (1) & \textbf{0.119}	& \textbf{90.0} &	\textbf{33.9}	& \textbf{54.3} & 	\textbf{71.3} \\
        
    \bottomrule
    \end{tabular}
    \vspace{-5pt}
    \label{tab:gt_supervised_training}
\end{table}

\subsection{Limitations}
As our method can be trained without any ground-truth annotations and scales naturally to large datasets, future work could explore training on larger and more diverse data to further improve generalization. Moreover, due to the absence of ground-truth supervision and the inherent limitations of 3DGS in recovering precise surface geometry~\cite{huang20242dgs}, additional finetuning with ground-truth depth is still beneficial for high-quality depth reconstruction.

\section{Conclusion}
In this paper we presente \ours, a self-supervised feed-forward approach that jointly predicts 3D Gaussian primitives and camera parameters from uncalibrated and unposed multi-view images. Extensive evaluations demonstrate state-of-the-art performance across multiple standard benchmarks, underscoring the potential of self-supervised models to make use of vast unlabeled data, and approach or even exceed the performance of their supervised counterparts in various 3D reconstruction tasks.

\clearpage

{
    \small
    \bibliographystyle{ieeenat_fullname}
    \bibliography{main}
}
\clearpage
\setcounter{page}{1}
\maketitlesupplementary

\section*{A. More Implementation Details}
\label{sec:appendix_details}
\textbf{More Architecture and Training Details.}
In our experiments, we apply NAS3R to two representative 3D models, MASt3R and VGGT. In Sec.~\ref{sec:results_no_priors}, NAS3R is trained from random initialization based on VGGT architecture, while in Sec.~\ref{sec:results_with_priors}, we train NAS3R on both VGGT and MASt3R architectures and utilize their respective pretrained weights for initialization.
Across all experiments, we use an initial learning rate of $1\times10^{-4}$ and set the LPIPS loss weight to 0.05. The batch size is 10 for VGGT-based models and 16 for MASt3R-based models. 

For the VGGT-based variant, we add a DPT-based Gaussian head while keeping other components as in the original VGGT: ViT-Large encoder (patch 14), alternating frame/global self-attention decoder, DPT depth head, and a camera head with self-attention layers and linear projection.
For the MASt3R-based variant, we use the original MASt3R ViT-Large encoder (patch 16) and extend the pairwise ViT-Base decoder to multi-view. We add DPT-based depth and Gaussian heads, and use a 3-layer MLP camera head that predicts rotation (6D), translation (4D homogeneous coordinates), and intrinsics (FOV).
For both variants, the decoder’s cross-attention is modified for masked context-to-target attention, and the camera head assumes shared intrinsics for all images within the same scene to ensure stable convergence.

\noindent\textbf{More Details on Baselines.}
All baselines use an input resolution of 256$\times$256, except those based on the VGGT architecture (SPFSplatV2-L and the NAS3R VGGT variant), which operate at 224$\times$224.
For the prior-free comparison in Sec.~\ref{sec:results_no_priors}, we retrain NoPoSplat and SPFSplatV2/V2-L with a 10k-step warm-up stage using the DUSt3R point-cloud distillation loss, following \cite{ye2025noposplat,huang2025spfsplat,huang2025spfsplatv2}.
In Tab. \ref{tab:pose_estimation}, SuperPoint + SuperGlue computes feature correspondences to estimate Essential Matrices and derive relative poses.
Since SelfSplat defines the target image as the reference frame, we evaluate relative poses between each pair of context images by setting the frist context image as the target image. For Fig.~\ref{fig:pose_traj}, the multi-view inference is decomposed to several two-view inference for SelfSplat.
In Tab. \ref{tab:pose_estimation_with_priors}, DUSt3R, MASt3R, and NoPoSplat estimate poses via PnP \cite{hartley2003multiple} with RANSAC~\cite{fischler1981ransac}, whereas PF3Splat, VGGT, and all SPFSplat variants directly regress camera poses.

\noindent\textbf{More Details on Downstream Finetuning.}
For Tab.~\ref{tab:gt_supervised_training}, following MapAnything~\cite{keetha2025mapanything}, we first compute the ground-truth pointmaps from the ground-truth poses and depth maps, as well as predicted pointmaps from the predicted poses and depth maps. Using the ground-truth validity masks, we compute scaling factors for both the ground-truth and up-to-scale predicted pointmaps, which are then used to normalize the depth maps and translations. The depth loss is adopted directly from MapAnything~\cite{keetha2025mapanything}.

For pose supervision, following~\cite{hong2024coponerf}, we combine a geodesic rotation loss and an $L_2$ translation loss. The rotation and translation terms are weighted by 0.1 and 0.01, respectively.

\noindent\textbf{More Details on Multi-View Experiments.}
For the multi-view experiments in Tab.~\ref{tab:multi_view_for_nvs}, we fix the first and last views and gradually increase the intermediate input views.

\section*{B. More Experimental Analysis}
\label{sec:appendix_experiments}

\begin{figure*}[t]
    \centering
    \includegraphics[width=1.0\textwidth]{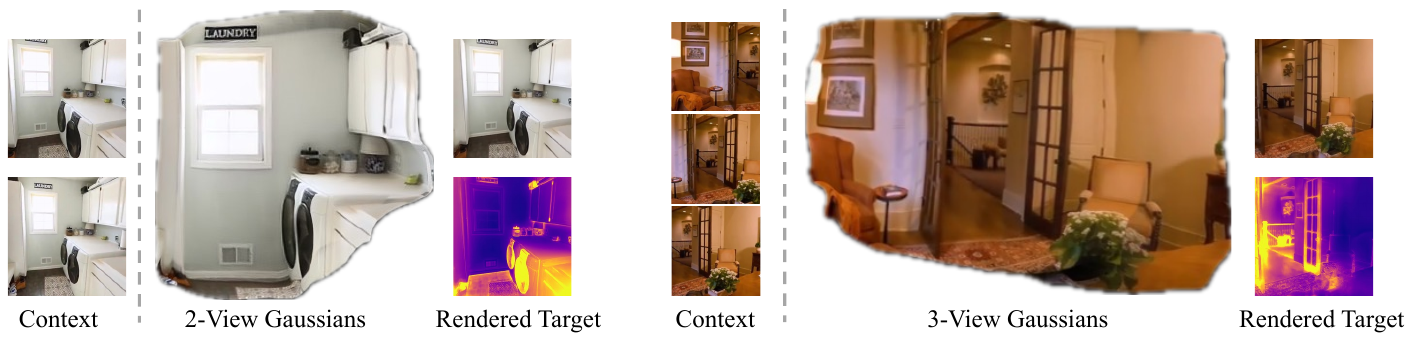}
    \vspace{-15pt}
    \caption{Examples of 3D Gaussians and rendered RGB and depth results on RE10K.}
    \label{fig:supp_gs_vis}
\end{figure*}
\begin{figure*}[t]
    \centering
    \includegraphics[width=0.99\textwidth]{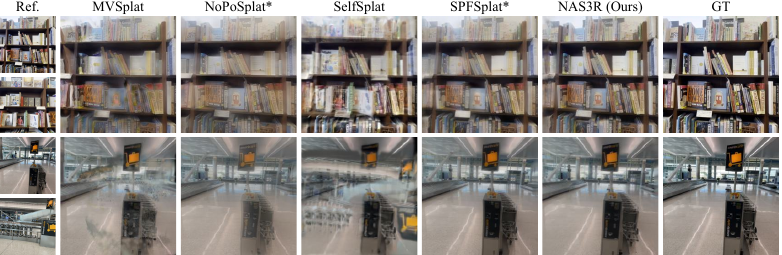}
    \vspace{-5pt}
    \caption{More novel view synthesis  qualitative comparisons on DL3DV.}
\label{fig:qualitative_comparison_dl3dv_sup}
\end{figure*}


\noindent\textbf{{Progressive Interval Curriculum.}}
We adopt a simple frame-interval-based strategy for view selection, a common practice in prior work~\cite{chen2024mvsplat, ye2025noposplat}, in which the interval between context frames is progressively increased during training. We also evaluate a \emph{no-curriculum} variant that samples from a fixed frame-interval range throughout training (Tab.~\ref{tab:no_curriculum}). Results show that, although performance is slightly lower without the curriculum, NAS3R still converges stably, demonstrating the robustness of our method.

\begin{table}[h]
\footnotesize
\setlength{\tabcolsep}{0.5pt}
\caption{Ablation on progressive interval curriculum on RE10K.}
\vspace{-5pt}
\centering
    \begin{tabular}{l ccc ccc}
        \toprule
        \multirow{2}{*}{\textbf{Method}} & \multicolumn{3}{c}{\textbf{NVS}} & \multicolumn{3}{c}{\textbf{Pose}} \\
        \cmidrule(lr){2-4} \cmidrule(lr){5-7}  
        & PSNR$\uparrow$ & SSIM$\uparrow$ & LPIPS$\downarrow$ &  5$^\circ$ $\uparrow$ & 10$^\circ$ $\uparrow$ & 20$^\circ$ $\uparrow$  \\
        \midrule
        No Curriculum & 21.020 & 0.650 & 0.255 & 20.3 & 35.8 & 50.2 \\
        Progressive Interval Curriculum  & \textbf{23.130}	& \textbf{0.764}	& \textbf{0.193} & \textbf{32.6}	& \textbf{51.0} & \textbf{64.9}\\

        \bottomrule
    \end{tabular}
\label{tab:no_curriculum}
\end{table}

\begin{table}[!t]
    \centering
    \footnotesize
    \caption{Ablation on ground-truth intrinsics on RE10K.}
    \vspace{-1em}
    \setlength{\tabcolsep}{3pt}

    \begin{tabular}{l c ccc ccc}
    \toprule
    \multirow{2}{*}{\textbf{Methods}} 
    & \multicolumn{3}{c}{\textbf{w/o priors}} 
    & \multicolumn{3}{c}{\textbf{w/ VGGT priors}} 
    \\
    \cmidrule(lr){2-4} \cmidrule(lr){5-7}
     & 
     PSNR $\uparrow$ & SSIM $\uparrow$ & LPIPS $\downarrow$ 
     & PSNR $\uparrow$ & SSIM $\uparrow$ & LPIPS $\downarrow$ 
      \\
    \midrule
     \ours & {23.130}	& {0.764}	& {0.193}  & {25.888}	& {0.861}	& {0.136}  \\
     \ours-I & 23.144 & 0.758 & 0.196 & 25.872	& 0.861 & 0.135\\

    \bottomrule
    \end{tabular}
    \vspace{-5pt}
    \label{tab:intrinsics_ablation}
\end{table}
\noindent\textbf{Ground-truth Intrinsics.}
Tab.~\ref{tab:pose_estimation_with_priors} show that ground-truth intrinsics improve pose estimation. Ablation results in Tab.~\ref{tab:intrinsics_ablation} indicates intrinsics have minimal impact on NVS quality, which is mainly because photometric supervision adjusts other 3D attributes to to be consistent with the self-learned intrinsics to achieve better rendering quality.

\noindent\textbf{{Comparison with AnySplat.}}
Although both of our method and AnySplat~\cite{jiang2025anysplat} adopt a local-to-global paradigm and VGGT-like architecture, there are two key differences.
(1) \textit{Depth Activation}: AnySplat follows the exponential depth activation used in VGGT, whereas our method predicts depth using a sigmoid activation followed by linear interpolation between near and far planes. This bounded formulation avoids an unbounded or explosive depth space, enabling stable training from random initialization.
(2) \textit{View Supervision}: AnySplat enforces the rendering loss only on the input context views, which leads overfitting to those views and frequent failures in depth and camera prediction. Consequently, it relies on pretrained VGGT weights and pseudo labels for stabilization.
In contrast, our method processes novel target views via masked attention and enforces photometric supervision on these unseen views, encouraging better generalization and stable training without any pretrained priors.
As shown in Tab.~\ref{tab:anysplat_comparison}, despite being trained with much less data and no pretrained priors, NAS3R outperforms AnySplat in the 2-view setting and achieves comparable performance with more views. 
\begin{table}[h]
\scriptsize
\setlength{\tabcolsep}{1pt}
\caption{Comparison with AnySplat on DL3DV dataset.}
    \centering
    \begin{tabular}{l l ccc ccc}
        \toprule
        \multirow{2}{*}{\textbf{Method}} & \multirow{2}{*}{\textbf{Training Data }}&\multicolumn{3}{c}{\textbf{2 View}} & \multicolumn{3}{c}{\textbf{5 Views}} \\
        \cmidrule(lr){3-5} \cmidrule(lr){6-8}  
        & & PSNR$\uparrow$ & SSIM$\uparrow$ & LPIPS$\downarrow$ & PSNR$\uparrow$ & SSIM$\uparrow$ & LPIPS$\downarrow$ \\
        \midrule
        (a) AnySplat & DL3DV+8 datasets & 16.905 & 0.514 & 0.249 & \textbf{21.680} & \textbf{0.694} & \textbf{0.218} \\
        (b) NAS3R & DL3DV & \textbf{20.069} & \textbf{0.588} & \textbf{0.281} & 21.307 & 0.646 & 0.240\\
        \bottomrule
    \end{tabular}
\label{tab:anysplat_comparison}
\end{table}

\noindent\textbf{Inference Efficiency.}
In Tab.~\ref{tab:inference_efficiency}, we report the inference efficiency of the VGGT-based and MASt3R-based NAS3R variants, measured on an A6000 GPU for reconstructing 3D Gaussians and predicting camera poses from two input images.


\begin{table}[ht]
\footnotesize
\caption{Inference efficiency on an NVIDIA A6000 GPU.}
\vspace{-5pt}
\setlength{\tabcolsep}{7.5pt}
\label{tab:inference_efficiency}
    \centering
    \begin{tabular}{lccc}
        \toprule
        Methods & Params (M) & FLOPs (GMac) & Time (s) \\
        \midrule
        
        NAS3R (MASt3R)   &  613.47 &  404.52 & 0.042 \\
        NAS3R (VGGT) & 1223.2 &  607.65 & 0.073 \\
        \bottomrule
    \end{tabular}
\end{table}

\section*{C. More Visualizations}
\noindent\textbf{Gaussian Reconstruction.}
Fig.~\ref{fig:supp_gs_vis} shows reconstructed Gaussians from two or three context views, along with the rendered RGB and depth maps,  illustrating our model’s ability to perform multi-view reconstruction and produce high-quality renderings.

\begin{figure*}[ht]
    \centering
    \includegraphics[width=0.99\textwidth]{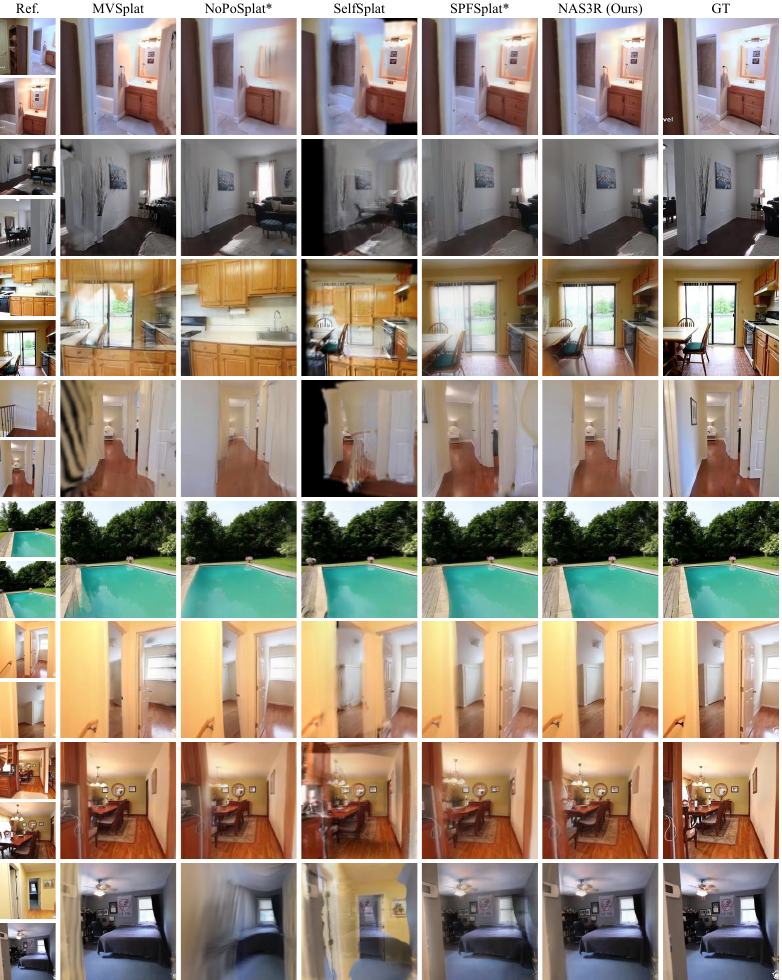}
    \caption{More novel view synthesis  qualitative comparisons on RE10K.}
\label{fig:qualitative_comparison_re10k_sup}
\end{figure*}

\begin{figure*}[ht]
    \centering
    \includegraphics[width=0.99\textwidth]{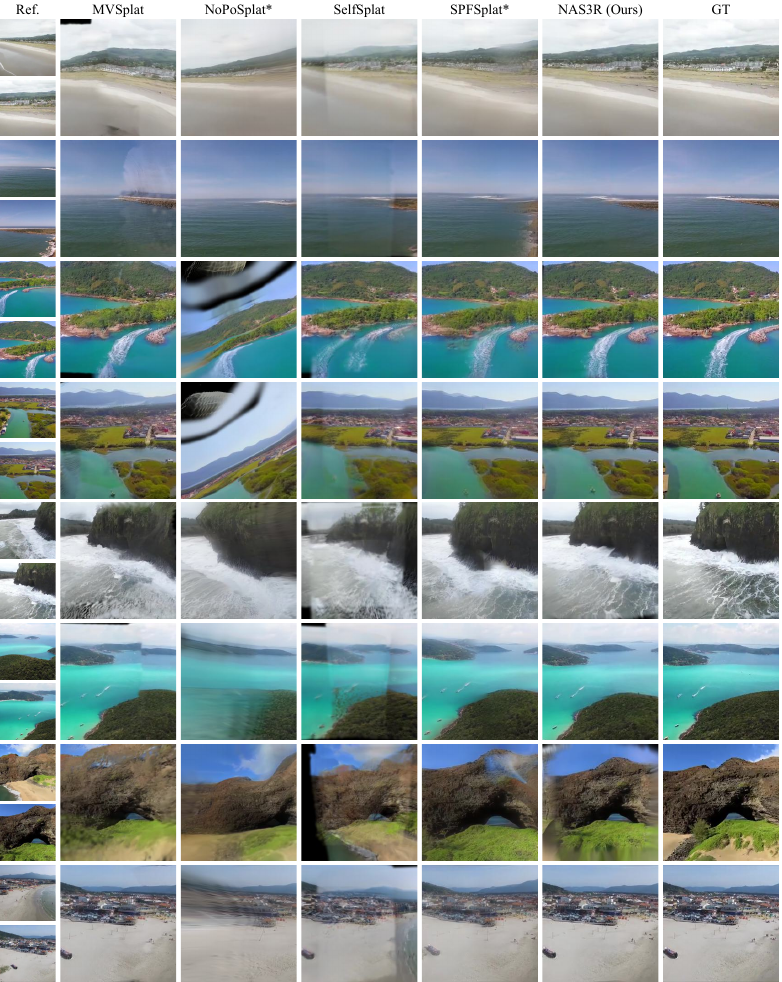}
    \caption{More novel view synthesis qualitative comparisons on ACID.}
\label{fig:qualitative_comparison_acid_sup}
\end{figure*}

\noindent\textbf{Novel View Synthesis.}
We provide additional NVS qualitative comparisons with baselines on DL3DV (Fig.~\ref{fig:qualitative_comparison_dl3dv_sup}), RE10K ( Fig.~\ref{fig:qualitative_comparison_re10k_sup}) and ACID ( Fig.~\ref{fig:qualitative_comparison_acid_sup}).
Across all cases, our method delivers strong NVS performance, even for image pairs with very small or no viewpoint overlap.

\end{document}